\documentclass[journal]{IEEEtai}

\usepackage{amsmath}
\usepackage{amssymb}

\usepackage{microtype}
\usepackage{graphicx}
\usepackage{subfigure}
\usepackage{booktabs} 
\usepackage{multirow}
\usepackage[table]{xcolor} 
\begin{document}
%
\title{RINS-T: Robust Implicit Neural Solvers for Time Series Linear Inverse Problems}
%
%
%

\author{Keivan~Faghih Niresi,
        Zepeng Zhang,
        and~Olga~Fink,~\IEEEmembership{Member,~IEEE}

\thanks{This research was supported by the Swiss Federal Institute of
Metrology (METAS).}
\thanks{Corresponding author: Olga Fink (olga.fink@epfl.ch)}
\thanks{Keivan Faghih Niresi, Zepeng Zhang, and Olga Fink are with the Intelligent Maintenance
and Operations Systems Laboratory, EPFL, 1015 Lausanne, Switzerland
(e-mail: keivan.faghihniresi@epfl.ch; zepeng.zhang@epfl.ch; olga.fink@epfl.ch).}
\thanks{Copyright (c) 2025 IEEE. Personal use of this material is permitted. However, permission to use this material for any other purposes must be obtained from the IEEE by sending a request to pubs-permissions@ieee.org}
}

%
%

\markboth{Submitted to IEEE Transactions on Instrumentation and Measurement
}%
{}
%



\maketitle

\begin{abstract}
Time series data are often affected by various forms of corruption, such as missing values, noise, and outliers, which pose significant challenges for tasks such as forecasting and anomaly detection. To address these issues, inverse problems focus on reconstructing the original signal from corrupted data by leveraging prior knowledge about its underlying structure. While deep learning methods have demonstrated potential in this domain, they often require extensive pretraining and struggle to generalize under distribution shifts. In this work, we propose RINS-T (Robust Implicit Neural Solvers for Time Series Linear Inverse Problems), a novel deep prior framework that achieves high recovery performance without requiring pretraining data. RINS-T leverages neural networks as implicit priors and integrates robust optimization techniques, making it resilient to outliers while relaxing the reliance on Gaussian noise assumptions. To further improve optimization stability and robustness, we introduce three key innovations: guided input initialization, input perturbation, and convex output combination techniques. Each of these contributions strengthens the framework's optimization stability and robustness. These advancements make RINS-T a flexible and effective solution for addressing complex real-world time series challenges. Our code is available at https://github.com/EPFL-IMOS/RINS-T.
\end{abstract}


\section{Introduction}
\begin{figure}
    \begin{center}
        \includegraphics[width=\linewidth]{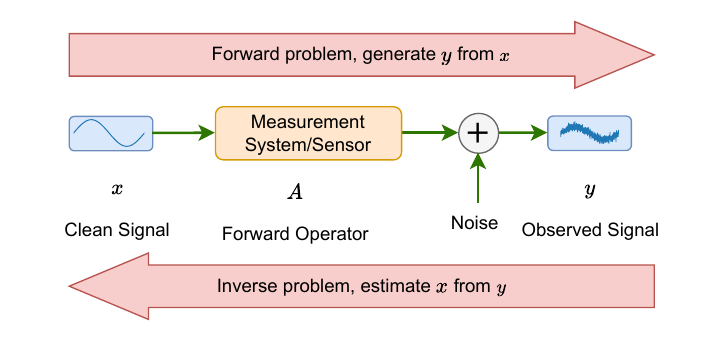}
    \end{center}
    \caption{Illustration of the measurement forward and inverse problems. In the forward problem, a clean signal passes through a measurement system (e.g., sensor and acquisition chain) and is degraded by noise, producing the observed signal. The inverse problem seeks to reconstruct the clean signal from these corrupted measurements.} 
    \label{inverseproblem}
\end{figure}

Time series data play a crucial role in a wide range of fields, including finance, healthcare, engineering, and environmental monitoring, as they capture temporal patterns and trends critical for decision-making and analysis. However, these datasets are often impacted by degradations such as missing values, noisy observations, and outliers, which can arise due to sensor failures, transmission errors, or environmental interference \cite{wen2023robust}. These degradations can severely compromise the performance of subsequent analyses and modeling tasks. Addressing these challenges is essential to ensure the integrity and utility of time series data. Many of these challenges can be formulated as inverse problems, where the goal is to reconstruct the clean signal from its corrupted or incomplete observations, ensuring that the reconstructed signal best explains the observed data while satisfying known constraints or priors regarding the signal's characteristics \cite{ongie2020deep}. From the perspective of instrumentation and measurement, these challenges naturally arise whenever physical sensor systems transform an underlying signal into noisy and degraded measurements \cite{zurita2020denoising, wang2021denoising, ou2024missing}. As shown in Fig. \ref{inverseproblem}, this process can be described as a forward problem in measurement science, where the sensor and acquisition chain act as a forward operator that maps the clean signal into its observed, noise-contaminated form. Solving the corresponding inverse problem (reconstructing the true signal from corrupted sensor data) is therefore central to maintaining the integrity and reliability of measurements. Traditional methods for addressing these challenges rely on a variety of approaches, including statistical techniques, signal processing, convex optimization, and machine learning algorithms \cite{alberti2021learning, shumaylov24a}. In optimization-based approaches to inverse problems, carefully defining data-fidelity term and prior or regularization term is a critical  step that directly influences solution quality and stability. Among these, the least-squares (LS) method is widely employed as a data-fitting term due to its mathematical simplicity and strong theoretical underpinnings. While  the LS method does not explicitly assume that errors follow an independent and identically distributed (i.i.d.) Gaussian distribution, it relies  on several important  assumptions: errors should have zero mean, constant variance (homoscedasticity), and be uncorrelated \cite{sun2021adjusting,gunther2022conditional}. According to the Gauss–Markov theorem, when these conditions are satisfied, the ordinary LS estimator is the best linear unbiased estimator, providing  the smallest variance among all linear unbiased estimators. However, the performance of the LS method deteriorates when the error distribution deviates from these assumptions, particularly in the presence of heavy-tailed distributions or heteroscedastic errors that are frequently encountered in real-world applications.  Its reliance on minimizing the sum of squared residuals also makes it highly sensitive to outliers, amplifying the influence of large deviations and leading to distorted estimates. This sensitivity to outliers is a critical limitation in real-world scenarios, where data contamination is common. Consequently, LS methods often struggle to manage outliers effectively, resulting in suboptimal performance in practical applications \cite{tukey1960survey, diakonikolas2022outlier}.
To address  these limitations, robust estimation techniques such as M-estimators have been proposed. M-estimators generalize the LS approach by minimizing alternative loss functions that reduce the influence of large deviations, thereby improving resilience to outliers \cite{muma2019robust}. Additionally, sparse recovery methods  are effective when the underlying signal can be represented sparsely  in a suitable  basis \cite{ito2019trainable, marques2018review}. However, their performance is highly sensitive to both the choice of basis and the degree of sparsity. If the true signal is not sufficiently sparse or the  assumed basis does not align well  with the actual sparsity structure, these methods may produce suboptimal or biased reconstructions. 

In addition to selecting an appropriate data-fitting term, formulating suitable priors or regularization terms is a fundamental aspect of model design. Priors incorporate domain knowledge to constrain the solution space, guiding the optimization process toward plausible and stable solutions. Traditionally, these priors are hand-crafted, such as smoothness constraints, sparsity, or total variation \cite{wen2018survey, unser2017splines, tropp2010computational}. More recently, neural networks have emerged as powerful implicit priors, with  the structure and inductive biases of the network itself imposing  useful constraints on the solution space. For instance, in  Deep Image Prior (DIP) \cite{ulyanov2018deep}, the network architecture acts as a prior, favoring natural-looking image reconstructions even when initialized with random weights and without any training data. By iteratively optimizing the network parameters to fit corrupted observations, DIP  enables tasks such as denoising \cite{niresi2022unsupervised}, inpainting \cite{niresi2023robust}, and super-resolution -- without  the need for pretraining or large datasets. Early implementations of DIP addressed the risk of  overparameterization   through strategies such as early stopping \cite{cheng2025effective, wangearly}. More recent approaches, however, have explored underparameterized models to improve computational efficiency and address tasks, such as image compression, denoising, super-resolution, and inpainting \cite{heckel2018deep}. Despite the proven success of unsupervised methods like DIP in image restoration, their application to time series data remains largely underexplored. A notable exception is  1D-DIP \cite{ravula2022one}, which demonstrated that 1D CNNs  can serve as effective implicit  priors for time series data, effectively addressing a range of  inverse problems including  denoising, imputation, and compressed sensing. While promising, the 1D-DIP approach has several limitations. The authors utilized the LS estimator for the data-fitting term and focused primarily on denoising performance under the assumption of ideal zero-mean Gaussian noise. Although theoretically sound, this setup does not fully reflect the complexities of real-world scenarios. Time series data are often corrupted by a variety of complex factors, including the presence of outliers, which are common in practical applications. Since the LS estimator is highly sensitive to outliers, its performance can degrade significantly in their presence. Moreover, many sensors used to  time series acquisition  operate within specific dynamic ranges, with   outputs constrained  by physical or electronic limits. In practice,  measurement noise can cause signal values near these boundaries  to become clipped or distorted. Such  non-linear effects are not captured by standard noise models, potentially leading to inaccuracies in downstream analysis. This clipping breaks the assumption of  Gaussian  noise \cite{foi2009clipped}, introducing additional complexities that the current 1D-DIP formulation does not address.

In this work, we address the critical limitations of deep prior methods for inverse problems in time series, especially under conditions where noise deviates from the Gaussian assumption. To overcome these challenges, we propose a robust framework that integrates multiple strategies designed to effectively handle contamination and outliers. Our main contributions are summarized as follows:

\begin{itemize}
\item We propose a novel framework for solving time series linear inverse problems under real-world degradations such as noise, outliers, and clipping, by leveraging deep prior architectures that do not rely on external training datasets.
\item We provide theoretical justification for employing the Huber loss function as a robust alternative to least squares in the presence of contaminated Gaussian noise, supported by two independent derivations that highlight its resilience against heavy-tailed distributions. 
\item We enhance the deep prior framework with a set of strategies, including guided input initialization, input perturbation, and convex output combination that collectively improve robustness, stability, and recovery performance. 
\item We demonstrate through extensive experiments on diverse time series datasets that the proposed framework consistently outperforms baseline methods, with particularly strong improvements in denoising, imputation, and compressed sensing under challenging noise and degradation conditions. 
\end{itemize}

\section{Related Works}

\textbf{Handcrafted Priors:} In many inverse problems, handcrafted priors are employed  to encode  structural assumptions about the signal, drawing   on domain knowledge or analytical convenience. These priors are explicitly designed to capture known data characteristics and are integrated  into optimization frameworks to guide the recovery process. For example, total variation (TV) regularization \cite{rudin1992nonlinear} is widely used in time series denoising tasks to preserve key transitions and structural features of the signal. In particular, TV regularization is effective for recovering signals with sharp changes or discontinuities, such as sudden trend shifts or abrupt variations in the data. Another widely  used prior is sparsity \cite{donoho2006compressed, candes2006robust}, which assumes that the signal can be represented with only a few non-zero coefficients in an suitable  transform domain, such as wavelets or Fourier bases  \cite{chang2000adaptive}. Sparse priors are particularly useful in scenarios where the underlying signal is sparse in the transformed domain, allowing effective recovery even from noisy or incomplete observations. While these handcrafted priors offer significant benefits, they depend heavily on prior assumptions about the data, which may not hold in real-world scenarios with complex noise or unexpected patterns. Additionally, combining multiple priors often leads to complex, non-convex optimization problems, which can increase computational costs and reduce scalability \cite{waters2011sparcs}. 

\textbf{Filtering Approaches:} In addition to regularization-based approaches,  a variety of  filtering techniques are commonly employed  for signal denoising \cite{mafi2019comprehensive}. Gaussian filtering smooths time series data by averaging local values, effectively suppressing  high-frequency noise while preserving general  trends. However, it can blur sharp transitions such as spikes or abrupt  shifts, and may perform poorly when the noise is non-Gaussian  or when the signal contains irregular fluctuations. Wiener filtering provides a more adaptive solution  by accounting  for both noise and signal variance, enabling  optimal smoothing that dynamically adjusts to the local signal-to-noise ratio. While powerful, its performance relies on accurate estimation of both the signal and noise characteristics, which may be challenging in real-world data with variable or unknown noise levels. Median filtering offers a non-linear approach by replacing each value in the signal with the median of its neighboring values. This approach is particularly effective for removing impulsive noise while preserving sharp features. However,  it may be less effective when  the noise is widespread  but not characterized by outliers, or in situations where preserving the inherent smoothness of the signal is especially important.

\textbf{Deep Learning Approaches:} Contemporary research increasingly applies  deep learning models  to time series inverse problems, leveraging techniques such as Generative Adversarial Networks (GANs) \cite{luo2018multivariate, yang2020adversarial, zhang2021missing}, Recurrent Neural Networks (RNNs) \cite{che2018recurrent, cao2018brits}, Autoencoders \cite{singh2022attention, langarica2023contrastive}, and denoising diffusion-based methods \cite{tashiro2021csdi}.  These approaches  are widely used  for tasks like denoising, signal recovery, and handling  incomplete or noisy data. However, a persistent  challenge for deep learning methods is their sensitivity to distributional shifts: when data characteristics or noise patterns  deviate from those seen during training, model  performance often declines. For instance, autoencoders and GANs may struggle to generalize to new types of noise or altered data patterns  if such variations are not well-represented in the training set.  To address  these limitations, recent work has explored integrating traditional signal processing  principles into deep learning architectures \cite{frusque2022learnable, michau2022fully}, leading to hybrid models  with improved interpretability, more meaningful filters, and reduced model complexity. Significant advances have been made  in both  image processing \cite{huang2021linn, huang2022winnet} and time series denoising \cite{frusque2024robust}, where incorporating signal processing techniques has enhanced generalizability. Nonetheless, these hybrid methods still depend on the availability of relevant pretraining data, restricting  their applicability in settings  where such data is limited or unavailable. An additional difficulty arises in univariate time series. Compared to multivariate time series, which provide richer information and can exploit structural relationships (e.g., graph representations combined with graph neural networks for recovery \cite{rey2022untrained, niresi2024physics} or downstream tasks \cite{li2024energy, li2025adaptive, niresi2025efficient}), univariate time series offer only a single signal, making them inherently more challenging to model and reconstruct. This lack of auxiliary information increases sensitivity to noise, outliers, and missing data, highlighting the importance of methods that can function effectively with minimal structural assumptions. Among deep learning-based approaches  for inverse problems, Plug-and-Play (PnP) algorithms have emerged as a flexible and effective class of approaches \cite{zhang2021missing}. Instead of explicitly defining a regularizer or prior, PnP methods integrate  powerful image or signal denoisers directly into iterative optimization schemes, enabling  practitioners to incorporate  state-of-the-art denoisers while preserving  a general optimization framework. Despite their success in imaging and signal restoration, PnP methods  face several limitations. Most notably, they typically rely on pre-trained denoisers tuned to specific noise levels or data distributions, which may not align with the actual characteristics of a given problem. Training  deep denoisers also requires access to large, domain-specific dataset, which can be difficult or impractical  to obtain. Furthermore, because the denoiser serves as implicit prior rather than an  explicit mathematical formulation, the underlying assumptions about the signal  are often opaque, complicating interpretation  and theoretical analysis. This lack of transparency makes it challenging to guarantee  reliable  performance outside the conditions seen during training. These limitations highlight the need for alternative approaches that combine the expressive capacity  of deep networks with stronger, problem-specific inductive biases and more principled theoretical foundations.  In contrast, deep prior approaches such as DIP \cite{ulyanov2018deep} and its extensions to time series \cite{ravula2022one} provide an unsupervised alternative. By leveraging the inherent  inductive bias of neural networks, these methods can perform effectively without pretraining, making them particularly valuable for  real-world scenarios  with scarce labeled data. Here, inductive bias refers to the set of assumptions encoded by neural network architectures about the underlying data structure and target problem, which enables them to generalize effectively from limited examples. Nevertheless, existing deep prior-based methods face difficulties  in dealing  with non-Gaussian noise and outliers, which are prevalent  in practical time series applications, highlighting the need for further advances in this area.

\section{Methodology}

In this section, we develop  the mathematical framework for signal reconstruction in the presence of  both Gaussian and sparse noise. We begin   by introducing key definitions to clearly outline  the problem setting. Next, we discuss  the concepts of forward and inverse problems, which set the stage for the presentation of our proposed RINS-T framework.

\textbf{Definition 3.1.} The \textit{infimal convolution} \cite{Bauschke2011convex} of two functions \( f \) and \( g \), where \( f, g: \mathbb{R}^N \to \mathbb{R} \cup \{+\infty\} \), is defined as:
\begin{equation}
    (f \square g)(x) := \inf_{v \in \mathbb{R}^N} \left\{ f(v) + g(x - v) \right\}.
\end{equation}
This operation generates a new function by combining \( f \) and \( g \) through a minimization process. The infimal convolution is particularly effective for \textit{smoothing} non-smooth convex functions, as it blends the behavior of \( f \) and \( g \) in a way that mitigates irregularities or discontinuities.

\textbf{Definition 3.2.} The \textit{Moreau envelope} \cite{moreau1965proximity, selesnick2017sparse} (also referred to as the Moreau-Yosida regularization) of a function \( f: \mathbb{R}^N \to \mathbb{R} \) is defined as:
\begin{equation}
    f^M(x) := \inf_{v \in \mathbb{R}^N} \left\{ f(v) + \frac{1}{2} \|x - v\|_2^2 \right\}.
\end{equation}

The Moreau envelope introduces \textit{smoothness} to the original function \( f \) by adding a quadratic regularization term that penalizes deviations between \( x \) and \( v \). This smoothing process retains essential properties of \( f \) while addressing its non-smooth behavior.

In the language of \textit{infimal convolution}, the Moreau envelope can be equivalently expressed as:
\begin{equation}
    f^M = f \square \frac{1}{2} \|\cdot\|_2^2.
\end{equation}
Here, the quadratic term \( \frac{1}{2} \|\cdot\|_2^2 \) acts as the smoothing function within the infimal convolution framework.

\textbf{Definition 3.3.} The \textit{proximal operator} \cite{parikh2014proximal} of a convex function \( f \) with parameter \( \lambda > 0 \) is defined as:
\begin{equation}
    \text{prox}_{\lambda f}(z) := \arg\min_{x \in \mathbb{R}^n} \left\{ \frac{1}{2} \|x - z\|_2^2 + \lambda f(x) \right\}.
\end{equation}
The proximal operator can be viewed as a regularized projection of \( z \) onto the set that minimizes \( f \). It balances proximity to \( z \) (via the squared Euclidean distance) with the minimization of \( f(x) \), controlled by the parameter \( \lambda \).

For the $\ell_1$-norm, the proximal operator corresponds to the \textit{soft-thresholding} operator, which promotes sparsity in the solution. Specifically, for $f(x) = \|x\|_1$, the proximal operator is given by:

\begin{equation}
\text{prox}_{\lambda \|\cdot\|_1}(z) =
\begin{cases}
z - \lambda & \text{if } z \geq \lambda, \\
0 & \text{if } -\lambda < z < \lambda, \\
z + \lambda & \text{if } z \leq -\lambda.
\end{cases}
\end{equation}

\subsection{Forward and Inverse Problems}

In many signal processing tasks, the observed signal \( y \in \mathbb{R}^m \) is a degraded version of the original clean signal \( x \in \mathbb{R}^n \), contaminated by both Gaussian noise \( g \sim \mathcal{N}(0, \sigma^2 I) \) and sparse noise \( s \in \mathbb{R}^m \). The forward model can be written as:
\begin{equation}
y = Ax + g + s.
\end{equation}

While the Gaussian noise term $g$ models small and continuous fluctuations such as sensor background noise or thermal variations, the sparse noise component $s$ captures infrequent disturbances that occur irregularly in real-world measurements. Such sparse noise may represent sudden spikes, signal clipping, or transient electrical interference. Modeling both Gaussian and sparse noise allows the framework to remain robust to a broader range of degradation patterns, improving reconstruction quality under realistic conditions. Given the observed signal \( y \) and known degradation operator $A \in \mathbb{R}^{m\times n}$, the goal is to recover \( x \) while identifying and mitigating the effect of sparse noise \( s \). This recovery task can be formulated as the following optimization problem:
\begin{equation}
\min_{x, s} \frac{1}{2} \| y - Ax - s \|_2^2 + \lambda \| s \|_1 + R(x),
\label{eq: inverseprob}
\end{equation}
where \( \frac{1}{2} \| y - Ax - s \|_2^2 \) is the data fidelity term accounting for Gaussian noise, \( \lambda \| s \|_1 \) promotes sparsity in the noise component \( s \), \( R(x) \) is a general regularization term (e.g., smoothness, sparsity, or other prior knowledge) imposed on the clean signal \( x \), and \( \lambda > 0 \) is a regularization parameter balancing the sparsity of \( s \) and the data fidelity term.

\subsection{Reformulation of the Optimization Problem}

The optimization problem involves two variables, \( x \) and \( s \), and is separable with respect to these variables. To solve it efficiently, we decompose the problem into two steps. The overall problem can be expressed as:
\begin{equation}
\min_{x} \min_{s} \left\{ \frac{1}{2} \| y - Ax - s \|_2^2 + \lambda \| s \|_1 + R(x) \right\}.
\end{equation}

We begin by solving the inner minimization problem with respect to \( s \), treating \( x \) as constant. The resulting optimization problem for \( s \) is:
\begin{equation}
\min_{s} \left\{ \frac{1}{2} \| y - Ax - s \|_2^2 + \lambda \| s \|_1 \right\}.
\label{eq:inner}
\end{equation}

To solve the minimization problem involving sparsity, we apply the \textit{proximal operator} of the scaled \( \ell_1 \)-norm. Let \( v = y - Ax \) be the residual between the observed signal \( y \) and the transformed clean signal \( Ax \). The optimization problem in \eqref{eq:inner} can then be rewritten as:
\begin{equation}
\min_{s} \left\{ \frac{1}{2} \| v - s \|_2^2 + \lambda \| s \|_1 \right\}.
\end{equation}

Since the problem is separable for each element of \( s \), we treat each component of \( s \) independently. Thus, for each \( s_i \), we solve the following scalar minimization problem:
\begin{equation}
\min_{s_i} \left\{ \lambda |s_i| + \frac{1}{2} (v_i - s_i)^2 \right\}.
\end{equation}

The solution \( s^* \) for each component \( s_i \) depends on the value of \( v_i \) relative to \( \lambda \). When \( |v_i| \leq \lambda \), the soft-thresholding operator gives \( s_i^* = 0 \). Substituting \( s_i^* = 0 \) into the objective function results in:
\begin{equation}
\lambda |0| + \frac{1}{2} (v_i - 0)^2 = \frac{1}{2} v_i^2.
\end{equation}

When \( v_i > \lambda \), the solution is \( s_i^* = v_i - \lambda \). Substituting this into the objective function gives:
\begin{equation}
\lambda |v_i - \lambda| + \frac{1}{2} \left( v_i - (v_i - \lambda) \right)^2 = \lambda (v_i - \lambda) + \frac{1}{2} \lambda^2.
\end{equation}
Simplifying the terms results in:
\begin{equation}
\lambda (v_i - \lambda) + \frac{1}{2} \lambda^2 = \lambda v_i - \frac{\lambda^2}{2}.
\end{equation}

When \( v_i < -\lambda \), the solution is \( s_i^* = v_i + \lambda \). Substituting this into the objective function gives:
\begin{equation}
\lambda |v_i + \lambda| + \frac{1}{2} \left( v_i - (v_i + \lambda) \right)^2 = \lambda (-v_i - \lambda) + \frac{1}{2} \lambda^2.
\end{equation}
Simplifying the terms results in:
\begin{equation}
\lambda (-v_i - \lambda) + \frac{1}{2} \lambda^2 = -\lambda v_i - \frac{\lambda^2}{2}.
\end{equation}

Combining the results from these cases, the optimal solution \( s^* \) leads to the following expression for the minimized objective function:
\begin{equation}
L_{\lambda}(v_i) =
\begin{cases}
\frac{1}{2} v_i^2, & \text{if } |v_i| \leq \lambda, \\
\lambda |v_i| - \frac{\lambda^2}{2}, & \text{if } |v_i| > \lambda.
\end{cases}
\end{equation}

\begin{figure}[h!]
  \centering
  \includegraphics[width=0.65\linewidth]{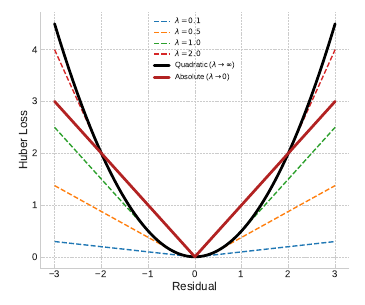}
  \caption{Illustration of the Huber loss transition as the threshold parameter $\lambda$ varies. Curves are shown for several representative $\lambda$ values.}
  \label{fig:huber_transition}
\end{figure}

This function \( L_{\lambda}(v) \) is the well-known Huber loss function \cite{huber1964robust}, which behaves quadratically for small residuals \( |v| \leq \lambda \) and transitions to a linear form for large residuals \( |v| > \lambda \). The parameter \( \lambda \) controls the threshold between these two regimes as shown in Fig. \ref{fig:huber_transition}.

The derived Huber loss function can be interpreted as the \textit{Moreau envelope} of the scaled \(\ell_1\)-norm \( \lambda \|\cdot\|_1 \). This connection highlights its role as a smooth approximation to the \(\ell_1\)-norm while retaining its sparsity-inducing properties.

\subsection{Probabilistic Interpretation}
\textbf{Definition 3.4.} A scaled version of the family \( P \) of \( \epsilon \)-\textit{contaminated Gaussian distributions} \cite{diakonikolas2024near} is given by:

\begin{equation}
P = \left\{ (1 - \epsilon)\Phi + \epsilon\Psi : \Psi \in S \right\},
\end{equation}
where \( 0 \leq \epsilon < \frac{1}{2} \), \( \Phi(t) \) is the standard cumulative distribution function of the inliers (Gaussian distribution), and \( S \) represents the set of the cumulative distributions of outliers. This model assumes that the degradation process consists of a mixture of two components: a fraction \( (1 - \epsilon) \) of Gaussian noise, and a fraction \( \epsilon \) of outliers. Huber \cite{huber1964robust} introduced the least favorable distribution
by modeling the degraded data as originating from an unknown distribution within the family $P$. 
The corresponding probability density function is expressed as: 

\begin{equation}
p_\lambda(t) = (1-\epsilon)\frac{1}{\sqrt{2\pi}} e^{-L_\lambda(t)},
\label{eq:LFD}
\end{equation}
where the parameter $\lambda$ is related to $\epsilon$ through (yielded from  $\int_{-\infty}^{\infty} p_\lambda(t)\, dt = 1$):  

\begin{equation}
\frac{\epsilon}{1-\epsilon} = \frac{2}{\lambda}\phi(\lambda) - 2\Phi(-\lambda),
\label{eq:rel}
\end{equation}
where $\phi(\lambda)$ is the standard normal probability density function.

We consider the observed signal \( y \in \mathbb{R}^n \) as a degraded version of the clean signal \( x \in \mathbb{R}^n \), contaminated by noise. The forward model is given by:

\begin{equation}
y = x + n,
\end{equation}
where \( n = g + s \) represents the noise, modeled as a contaminated Gaussian distribution (mixture of Gaussian and Laplace components). The noise distribution is governed by the magnitude of the residual \( |y - x| \): Gaussian noise predominates when the residual is small (\( |y - x| \leq \lambda \)), while Laplace noise becomes dominant for larger residuals (\( |y - x| > \lambda \)) \cite{meyer2021alternative}. To estimate \( x \), we adopt the Maximum A Posteriori (MAP) framework:

\begin{equation}
\begin{split}
\hat{x} &= \arg\max_{x} \prod_{i=1}^n p(x_i | y_i) \\
&= \arg\min_{x} \left( -\sum_{i=1}^n \log p(x_i | y_i) \right).
\end{split}
\end{equation}




By applying Bayes' rule and assuming the noise is independently drawn from either a Gaussian or Laplace distribution, the MAP estimate is derived by maximizing \( p(x_i|y_i) \). This is  equivalent to minimizing the negative log-posterior:

\begin{equation}
\begin{split}
-\sum_{i=1}^n\log p(x_i|y_i) 
&= -\sum_{i=1}^n\log p(y_i|x_i) \\
&\qquad - \sum_{i=1}^n\log p(x_i) + \text{const}.
\end{split}
\end{equation}

Here, \( p(y_i|x_i) \) is the likelihood of observing \( y_i \) given \( x_i \), \( p(x_i) \) is the prior distribution of \( x_i \), \( -\log p(x_i) \) acts as a regularization term \( R(x) \), incorporating prior knowledge about \( x \). The remaining term, \( -\log p(y_i|x_i) \), is determined by the noise model with the following relation:

\begin{equation}
p(y_i|x_i) \propto
\begin{cases}
\exp\left( -\frac{(y_i - x_i)^2}{2} \right) & \text{if } |y_i - x_i| \leq \lambda, \\
\exp\left( -\lambda |y_i - x_i| + \frac{\lambda^2}{2} \right) & \text{if } |y_i - x_i| > \lambda.
\end{cases}
\end{equation}

The negative log-likelihood then corresponds to the Huber loss function:

\begin{equation}
-\log p(y_i|x_i) = L_{\lambda}(y_i - x_i).
\end{equation}

By including the prior on \( x \) into the model, the MAP estimate introduces  a regularization term \( R(x) \), resulting  in the final objective:

\begin{equation}
\hat{x} = \arg \min_x \left\{ \sum_i L_{\lambda}(y_i - x_i) + R(x) \right\}.
\end{equation}

This formulation provides a robust estimator for \( x \) by leveraging the Huber loss to effectively address mixed Gaussian-Laplace noise while integrating prior knowledge through a regularization term.

\subsection{Robust Implicit Neural Solvers for Time Series Inverse Problems}
We introduce a robust framework that leverages the flexibility of deep priors while overcoming their limitations in handling noisy and outlier-contaminated time series data. To ensure improved recovery performance, we propose to extend our approach with robust fitting, guided input initialization, input perturbation, and convexly combined output.

\textbf{Robust Fitting:} To effectively solve inverse problems, it is crucial to complement a robust data-fitting term with an appropriate prior or regularization term $R(x)$. Recent advances have demonstrated that CNNs can generate high-quality, natural-looking images even when initialized with random weights and without any  pretraining on large datasets \cite{ulyanov2018deep}. This observation  challenges traditional approaches  that rely on extensive training, and highlights the intrinsic  structural properties of neural networks, which implicitly encode powerful priors. Given the conceptual similarities  between image and time series priors -- such as the need for smoothness, sparsity, or other structured constraints -- the deep prior framework presents a promising solution  for time series inverse problems. In this  framework, the architectural bias of the network is exploited by optimizing its weights while keeping the latent input vector fixed, enabling the model to adapt to the specific structure of the observed data. This approach  is  especially useful  when  it is difficult to  explicitly formulate suitable   priors. The optimization problem  can be formalized as follows:

\begin{equation}
\theta^* = \arg\min_\theta L_\lambda( y - Af_\theta(z)),
\end{equation}

where \( f_\theta(z) = \hat{x} \) is the output of the CNN given a fixed randomly initialized input \( z \) and trainable weights \( \theta \). The network weights \( \theta \) are iteratively updated during the optimization process to minimize the reconstruction loss, defined as the discrepancy between the observed data \( y \) and the CNN output \( \hat{x} \). This process exploits the inherent inductive bias of the CNN architecture, which naturally favors solutions that are structured and coherent, effectively acting as an implicit prior.

\textbf{Guided Input:} In standard deep prior frameworks, the input $z$ is typically initialized as random noise, requiring the network to learn a mapping from an entirely unstructured input to the desired output. This  makes optimization more challenging and increases the risk of overfitting high-frequency noise due to the network’s inherent spectral bias. To address this, we propose a guided input strategy: instead of random noise, $z$ is initialized with a smoothed version of the corrupted observation $y$, denoted as $u$, obtained via Gaussian filtering. This approach  suppresses high-frequency noise while preserving the underlying structure  of the time series, giving  the network  a more informative starting point. The benefits aretwofold. First, from a theoretical perspective, Neural Tangent Kernel (NTK) theory  suggests that the network’s  output at each  iteration is closely tied to the initial output, which  depends on  $z$, particularly  in architectures with skip connections \cite{tachella2021neural, alkhouri2024image}. In overparameterized networks, where  the Jacobian remains nearly  constant, the optimization trajectory  is largely determined by the input. Starting  with a structured signal like $u$ leads to more stable and meaningful convergence. Second, from a frequency-domain standpoint, this initialization reduces the network’s tendency to overfit high-frequency noise. Prior studies have shown that deep networks tend to learn low-frequency components first (spectral bias), and the frequency content of the input influences  what the network learns \cite{rahaman2019spectral}. While approaches such as Neural Radiance Fields (NeRF) \cite{mildenhall2021nerf}, use  high-frequency input embedding  to capture fine details that would otherwise be missed due to spectral bias \cite{liu2024architecture}, our method takes the complementary approach:  smoothing the input to  discourage noise fitting  and regularize the reconstruction process. This allows  the network to focus on refining relevant  details rather than reconstructing the signal from scratch. Our results demonstrate   that guided input improves convergence speed, enhances robustness, and consistently yields higher-quality reconstructions. These findings highlight guided input as a simple yet powerful improvement  to the deep prior framework.

\begin{figure*}
    \begin{center}
        \includegraphics[width=\linewidth]{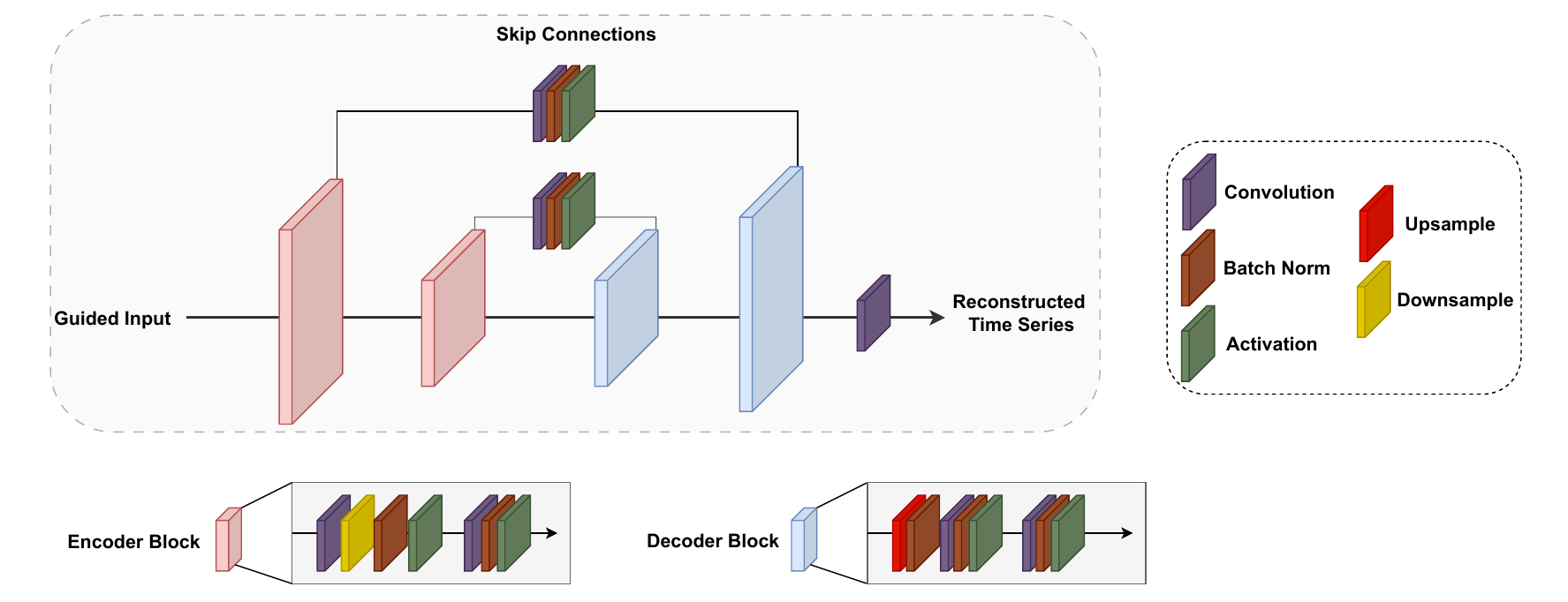}
    \end{center}
    \caption{Overview of the architecture used for Deep Prior, including encoder and decoder blocks connected through skip connections.} 
    \label{architecture}
\end{figure*}

\textbf{Input Perturbation:} To further enhance robustness, we introduce an input perturbation strategy inspired by the jittering technique from \cite{krainovic2023learning}. At each optimization step, Gaussian noise is added to the guided input, resulting in a perturbed input \( z_t = u + \epsilon_t \), where \( \epsilon_t \sim \mathcal{N}(0, \sigma^2) \) and \( u \) is the smoothed version of the corrupted observation. The optimization objective becomes  minimizing the expected loss over these perturbations:
\begin{equation}
\theta^* = \arg\min_\theta \mathbb{E}_{\epsilon_t \sim \mathcal{N}(0, \sigma^2)} \left[ L_\lambda(y - Af_\theta(z_t)) \right].
\end{equation}
While conceptually related  to input jittering, where noise is injected into an already noisy input and the network is trained to recover the clean target, our approach  differs by applying perturbations to a smoothed input , and  operating entirely in an unsupervised manner, without access to ground truth. This encourages the network to learn features that are invariant  to small variations, thereby  regularizing the learning  process. By exposing the model to multiple noisy realizations of the input, we reduce overfitting to specific features  of the guided input \( u \) and promote the extraction of consistent, generalizable structures. As a result, the model achieves better reconstruction quality and robustness against  noise and artifacts.

\textbf{Convexly Combined Output:} To promote  optimization stability and faster  convergence,  we update the model's output at each step using a convex combination: 
\begin{equation} 
f_{\theta_{t}}(z_t) \leftarrow \alpha \cdot f_{\theta_{t-1}}(z_{t-1}) + (1 - \alpha) \cdot f_{\theta_{t}}(z_t),
\end{equation} 
where \( \alpha \) is a weighting factor that controls  the contribution  of the previous  and current outputs.  This technique stabilizes training  by smoothly blending  current predictions with historical outputs, effectively reducing fluctuations and suppressing noise. The resulting soft regularization effect mitigates oscillations and prevents abrupt changes across iterations, leading to more reliable and robust convergence.

\subsection{Architectural Design Considerations for Deep Prior} We employ a hierarchical CNN specifically designed for one-dimensional data, such as time series. Inspired by the U-Net architecture, our model employs encoders, decoders, and skip connections.  The hierarchical structure consists of  multiple convolutional layers with varying kernel sizes and strides, enabling multi-resolution feature  extraction. Downsampling is performed via  strided convolutions, while skip connections integrate  low-level and high-level representations, preserving fine  details and global context simultaneously. In the final stages, the aggregated  features are mapped to a single output channel through a one-dimensional convolution, making the architecture well suited for tasks such as univariate time series reconstruction. An shown  of network is provided in Figure \ref{architecture}, and all hyperparameters are summarized in Table \ref{tab:hyperpara}.

\begin{table}[h!]
\centering  
\caption{Additional Hyperparameters}
\label{tab:hyperpara}
\begin{sc}  
\begin{tabular}{l l}
\toprule
\textbf{Hyperparameter} & \textbf{Value} \\
\midrule
Number of Encoder Layers & 2 \\
Number of Decoder Layers & 2 \\
Number of Skip Layers & 2 \\
Encoder Channel Sizes & [64, 64] \\
Decoder Channel Sizes & [64, 64] \\
Skip Channel Sizes & [4, 4] \\
Encoder Kernel Size& 3 \\
Decoder Kernel Size & 3 \\
Skip Kernel Size & 1 \\
Activation Function & LeakyReLU \\
Upsample Mode & Nearest \\
Downsample Mode & Stride \\
Optimizer & Adam \\
Learning Rate & 0.01 \\
$\alpha$ (Convex Combination) & 0.5 \\
$\lambda$ (Huber Loss Function) & 0.001 \\
\bottomrule
\end{tabular}
\end{sc}
\end{table}

\begin{figure}
    \centering
    \includegraphics[width=\linewidth]{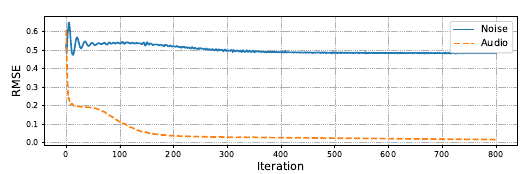}
    \caption{Ability of the Network to Fit Gaussian Noise and a Clean Audio Signal}
    \label{fig:bias}
\end{figure}

\textbf{Architectural Inductive Bias:} To characterize  the inductive bias of our deep prior architecture, we assess  its capacity  to fit both a clean audio signal and random noise, each initialized with random weights. As illustrated  in Figure \ref{fig:bias}, the network rapidly fits the structured audio signal, whereas  fitting random noise is considerably slower and less efficient. This contrast  underscores  the architecture's   inherent preference for learning structured patterns over unstructured noise -- a property that can be effectively exploited in time series inverse problems where the underlying signals possess meaningful structure.

\section{Experimental Results}

\subsection{Datasets}
We evaluated our approach on three publicly available datasets, each representing   a distinct type of univariate time series:

\begin{itemize}

\item \textbf{Audio}: Recordings of the Atlantic Spotted Dolphin vocalizations  \cite{knapp}. This time series exhibits non-stationary behavior due to changes in frequency and amplitude over time, as is typical in animal acoustic signals.

\item \textbf{Electricity}: Hourly  electricity consumption data  (kWh), where extracted the time series for the second client to obtain a univariate sequence \cite{electricityloaddiagrams20112014_321}. The data exhibits periodic or seasonal patterns, such as daily or weekly consumption cycles.

\item \textbf{Solar}: Solar power production data from 2006, sampled every 10 minutes from 137 photovoltaic (PV) plants in Alabama. We used the time series from the second PV plant to form a univariate dataset\footnote{https://www.nrel.gov/grid/solar-power-data.html}. Solar data often contains daily trends and can be affected by weather variability, which may result in non-stationary and partially periodic behavior.

\end{itemize}

All datasets were preprocessed with min-max normalization. For quantitative evaluation, we report  Root Mean Squared Error (RMSE), Mean Absolute Error (MAE), and Signal-to-Noise Ratio (SNR).

\begin{table*}[t]
    \centering
    \caption{Comparison of Different Denoising Methods across 3 Datasets}
    \resizebox{\linewidth}{!}{
    \begin{sc}
    \begin{tabular}{llcccccccccccc}
        \toprule
        \multirow{2}{*}{Scenarios} & \multirow{2}{*}{Method} & \multicolumn{3}{c}{Audio} & \multicolumn{3}{c}{Electricity} & \multicolumn{3}{c}{Solar} & \multicolumn{3}{c}{\cellcolor{gray!20}Average} \\
        \cmidrule(lr){3-5} \cmidrule(lr){6-8} \cmidrule(lr){9-11} \cmidrule(lr){12-14}
        & & RMSE $\downarrow$ & MAE $\downarrow$ & SNR $\uparrow$ & RMSE $\downarrow$ & MAE $\downarrow$ & SNR $\uparrow$ & RMSE $\downarrow$ & MAE $\downarrow$ & SNR $\uparrow$ & \cellcolor{gray!20}RMSE $\downarrow$ & \cellcolor{gray!20}MAE $\downarrow$ & \cellcolor{gray!20}SNR $\uparrow$ \\
        \midrule
        \multirow{8}{*}{Scenario 1} & Noisy & 0.0988 & 0.0790 & 13.78 & 0.0997 & 0.0795 & 11.34 & 0.0819 & 0.0543 & 11.79 & \cellcolor{gray!20}0.0935 & \cellcolor{gray!20}0.0709 & \cellcolor{gray!20}12.30 \\
        & Gaussian & 0.0392 & 0.0315 & 21.81 & 0.0647 & 0.0504 & 15.09 & 0.0518 & 0.0432 & 15.77 & \cellcolor{gray!20}0.0519 & \cellcolor{gray!20}0.0417 & \cellcolor{gray!20}17.56 \\
        & Median & 0.0735 & 0.0590 & 16.35 & 0.0706 & 0.0556 & 14.34 & 0.0556 & 0.0369 & 15.15 & \cellcolor{gray!20}0.0666 & \cellcolor{gray!20}0.0505 & \cellcolor{gray!20}15.28 \\
        & Wiener & 0.0582 & 0.0455 & 18.38 & \underline{0.0582} & \underline{0.0453} & \underline{16.01} & 0.0556 & 0.0437 & 15.16 & \cellcolor{gray!20}0.0573 & \cellcolor{gray!20}0.0448 & \cellcolor{gray!20}16.52 \\
        & Wavelet & \underline{0.0365} & \underline{0.0294} & \underline{22.42} & 0.0624 & 0.0491 & 15.41 & 0.0495 & \underline{0.0409} & 16.16 & \cellcolor{gray!20}0.0495 & \cellcolor{gray!20}0.0398 & \cellcolor{gray!20}17.99 \\
        & TV & \textbf{0.0359} & \textbf{0.0290} & \textbf{22.56} & \textbf{0.0548} & \textbf{0.0430} & \textbf{16.54} & \underline{0.0486} & 0.0417 & \underline{16.31} & \cellcolor{gray!20}\underline{0.0464} & \cellcolor{gray!20}\underline{0.0379} & \cellcolor{gray!20}\underline{18.47} \\
        & 1D-DIP & 0.0675 & 0.0538 & 17.09 & 0.0714 & 0.0562 & 14.24 & 0.0606 & 0.0486 & 14.40 & \cellcolor{gray!20}0.0665 & \cellcolor{gray!20}0.0529 & \cellcolor{gray!20}15.24 \\
        & RINS-T & 0.0389 & 0.0312 & 21.87 & 0.0616 & 0.0468 & 15.53 & \textbf{0.0364} & \textbf{0.0208} & \textbf{18.84} & \cellcolor{gray!20}\textbf{0.0456} & \cellcolor{gray!20}\textbf{0.0329} & \cellcolor{gray!20}\textbf{18.75} \\
        \midrule
        \multirow{8}{*}{Scenario 2} & Noisy & 0.2586 & 0.2110 & 5.42 & 0.2633 & 0.2169 & 2.91 & 0.2273 & 0.1501 & 2.92 & \cellcolor{gray!20}0.2497 & \cellcolor{gray!20}0.1927 & \cellcolor{gray!20}3.75 \\
        & Gaussian & 0.0776 & 0.0623 & 15.87 & 0.0941 & 0.0738 & 11.85 & \underline{0.1188} & \underline{0.1050} & \underline{8.55} & \cellcolor{gray!20}0.0968 & \cellcolor{gray!20}0.0804 & \cellcolor{gray!20}12.09 \\
        & Median & 0.1881 & 0.1519 & 8.19 & 0.1918 & 0.1546 & 5.66 & 0.1552 & 0.1032 & 6.24 & \cellcolor{gray!20}0.1784 & \cellcolor{gray!20}0.1366 & \cellcolor{gray!20}6.70 \\
        & Wiener & 0.1419 & 0.1108 & 10.64 & 0.1419 & 0.1119 & 8.28 & 0.1557 & 0.1199 & 6.21 & \cellcolor{gray!20}0.1465 & \cellcolor{gray!20}0.1142 & \cellcolor{gray!20}8.38 \\
        & Wavelet & \textbf{0.0547} & \textbf{0.0439} & \textbf{18.92} & \underline{0.0917} & \underline{0.0703} & \underline{12.07} & 0.1236 & 0.1060 & 8.21 & \cellcolor{gray!20}\underline{0.0900} & \cellcolor{gray!20}\underline{0.0734} & \cellcolor{gray!20}\underline{13.07} \\
        & TV & 0.0765 & 0.0611 & 16.00 & \textbf{0.0867} & \textbf{0.0690} & \textbf{12.56} & 0.1216 & 0.1100 & 8.35 & \cellcolor{gray!20}0.0949 & \cellcolor{gray!20}0.0800 & \cellcolor{gray!20}12.30 \\
        & 1D-DIP &  0.1579 & 0.1259 & 9.71 & 0.0942 & 0.0724 & 11.83 & 0.1569 & 0.1267 & 6.14 & \cellcolor{gray!20}0.1363 & \cellcolor{gray!20}0.1083 & \cellcolor{gray!20}9.23 \\
        & RINS-T & \underline{0.0717} & \underline{0.0576} & \underline{16.56} & 0.0922 & 0.0738 & 12.02 & \textbf{0.0765} & \textbf{0.0532} & \textbf{12.38} & \cellcolor{gray!20}\textbf{0.0801} & \cellcolor{gray!20}\textbf{0.0615} & \cellcolor{gray!20}\textbf{13.65} \\
        \midrule
        \multirow{8}{*}{Scenario 3} & Noisy & 0.1480 & 0.1015 & 10.27 & 0.1441 & 0.1001 & 8.14 & 0.1911 & 0.1164 & 4.43 & \cellcolor{gray!20}0.1611 & \cellcolor{gray!20}0.1060 & \cellcolor{gray!20}7.61 \\
        & Gaussian & 0.0525 & 0.0412 & 19.27 & 0.0732 & 0.0579 & 14.02 & \underline{0.0764} & \underline{0.0591} & \underline{12.39} & \cellcolor{gray!20}0.0674 & \cellcolor{gray!20}0.0527 & \cellcolor{gray!20}15.23 \\
        & Median & 0.0876 & 0.0668 & 14.83 & 0.0868 & 0.0642 & 12.55 & 0.0989 & 0.0669 & 10.15 & \cellcolor{gray!20}0.0911 & \cellcolor{gray!20}0.0660 & \cellcolor{gray!20}12.51 \\
        & Wiener & 0.0995 & 0.0616 & 13.72 & 0.0960 & 0.0609 & 11.67 & 0.1376 & 0.0724 & 7.28 & \cellcolor{gray!20}0.1110 & \cellcolor{gray!20}0.0650 & \cellcolor{gray!20}10.89 \\
        & Wavelet & \textbf{0.0442} & \textbf{0.0352} & \textbf{20.76} & 0.0772 & 0.0595 & 13.56 & 0.0863 & 0.0636 & 11.33 & \cellcolor{gray!20}0.0692 & \cellcolor{gray!20}0.0528 & \cellcolor{gray!20}15.22 \\
        & TV & \underline{0.0460} & \underline{0.0359} & \underline{20.43} & \underline{0.0728} & \underline{0.0558} & \underline{14.07} & 0.0791 & 0.0617 & 12.09 & \cellcolor{gray!20}\underline{0.0660} & \cellcolor{gray!20}\underline{0.0511} & \cellcolor{gray!20}\underline{15.53} \\
        & 1D-DIP & 0.0978 & 0.0753 & 13.87 & 0.0903 & 0.0700 & 12.20 & 0.1125 & 0.0820 & 9.03 & \cellcolor{gray!20}0.1002 & \cellcolor{gray!20}0.0758 & \cellcolor{gray!20}11.70 \\
        & RINS-T & 0.0480 & 0.0380 & 20.05 & \textbf{0.0698} & \textbf{0.0536} & \textbf{14.44} & \textbf{0.0526} & \textbf{0.0311} & \textbf{15.64} & \cellcolor{gray!20}\textbf{0.0568} & \cellcolor{gray!20}\textbf{0.0409} & \cellcolor{gray!20}\textbf{16.71} \\
        \bottomrule
    \end{tabular}
    \end{sc}
    }
    \label{tab:denoising}
\end{table*}

\subsection{Denoising}

In the denoising experiments, the sensing matrix $A$ is set to the identity matrix $I$,  meaning the observed signals are direct but noisy versions of the original signal.

\textbf{Baselines:} We compared RINS-T to several established denoising methods: (1) Gaussian filtering \cite{deisenroth2011general}, (2) Median filtering \cite{yin1996weighted}
, (3) Wiener filtering \cite{oppenheim2018signals}, (4) Wavelet denoising using sym4 wavelets \cite{daubechies2002wavelet}, (5) Total Variation (TV) denoising \cite{rudin1992nonlinear}, and (6) 1D-DIP \cite{ravula2022one}. In addition, we include the raw noisy signals (``Noisy”) as a reference to illustrate the effect of denoising. For a fair comparison, both 1D-DIP and RINS-T use the same untrained CNN architecture, and we report the average results across five independent trials. Although Gaussian, Median, and Wiener filtering all belong to the class of filtering methods, they have complementary properties: Gaussian filtering smooths noise, Median filtering effectively removes impulsive noise while preserving sharp changes, and Wiener filtering adaptively minimizes mean-square error. Including all three allows a comprehensive comparison of filtering approaches.

\textbf{Noise Scenarios:} To assess robustness across different noise conditions, we considered  three scenarios designed to reflect both common and challenging types of corruption encountered in real-world time series data. These include mild and severe Gaussian noise, as well as the presence of outliers, which are particularly problematic for standard estimation techniques:

\begin{itemize}

\item \textbf{Scenario  1}: Zero-mean Gaussian noise with standard deviation 0.1 was added to the normalized data. The resulting  signal was  clipped to the  $[0, 1]$ range, introducing deviations from the ideal Gaussian distribution.

\item \textbf{Scenario  2}: Gaussian noise with a higher  a standard deviation of 0.3 was added, followed by clipping. This scenario produced stronger distortion and more significant deviations from the original signal.

\item \textbf{Scenario  3}: Gaussian noise with a standard deviation of 0.1 was combined  with 10\% outliers,  where outlier magnitudes were uniformly sampled from $[0, 1]$. No clipping was applied in this scenario.
\end{itemize}

\textbf{Results:} Table \ref{tab:denoising} presents  a comprehensive  comparison of  denoising methods across nine different scenarios (three noise scenarios applied to three datasets). RINS-T achieves the best results in four out of nine scenarios and delivers the highest average performance across all metrics. TV Denoising and Wavelet Denoising also perform well, with TV achieving top results in three scenarios and Wavelet in two reflecting their strengths for specific types of noise and data. RINS-T performs particularly well on the Electricity and Solar datasets, as  these data exhibit smooth and  structured temporal dynamics -- such as periodicity and long-term trends -- that align well with the architectural inductive biases of  RINS-T's untrained neural network. The method also demonstrates  resilience in Scenario 3, where outliers are present, due  to its robust data-fitting term and strategies such as guided input perturbation and convex output averaging. 

\begin{figure}
    \begin{center}
        \includegraphics[width=\linewidth]{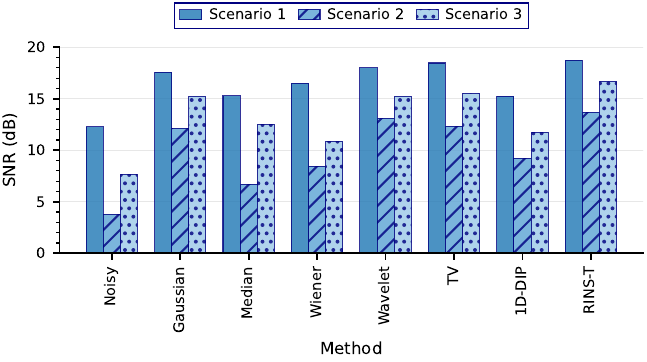}
    \end{center}
    \caption{Comparison of average SNR values across three denoising scenarios using different methods. The proposed RINS-T method consistently achieves the best SNR across all scenarios.} 
    \label{denoising_chart}
\end{figure}

\begin{figure*}
  \centering
\includegraphics[scale=0.4]{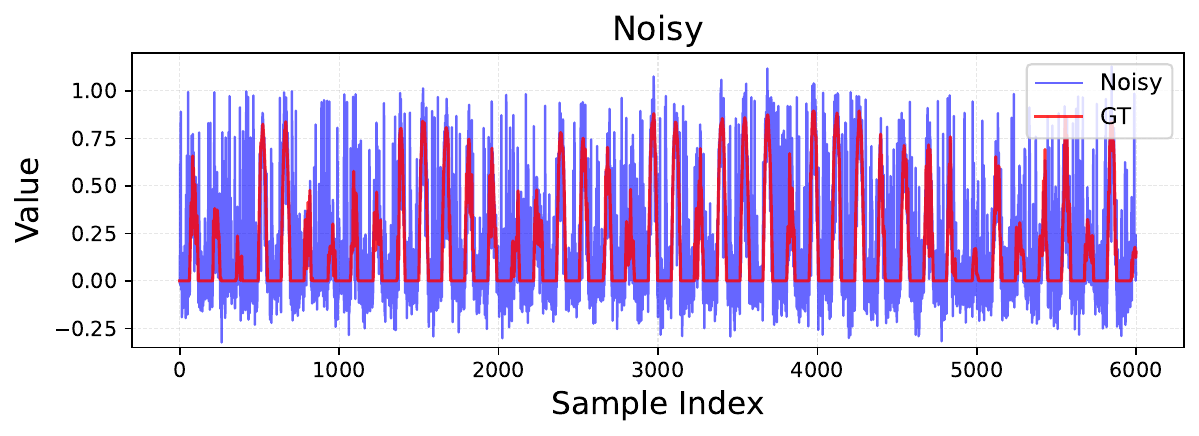}\hspace{0.0em}%
\includegraphics[scale=0.4]{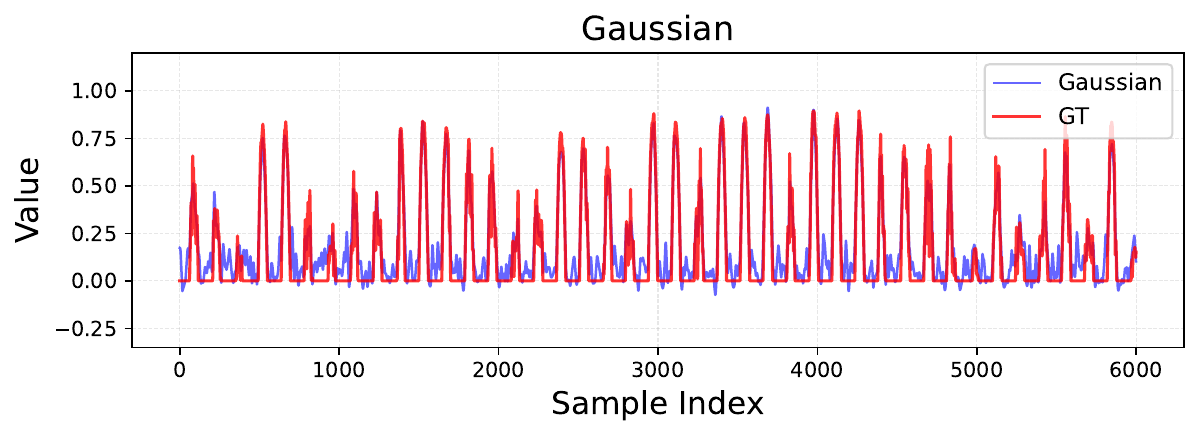}\hspace{0.0em}
\includegraphics[scale=0.4]{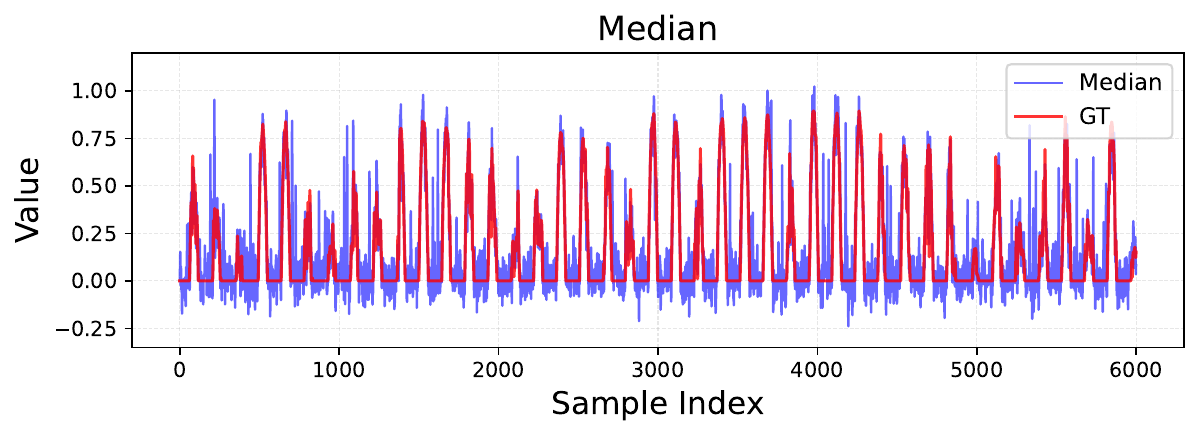}\hspace{0.0em}
\includegraphics[scale=0.4]{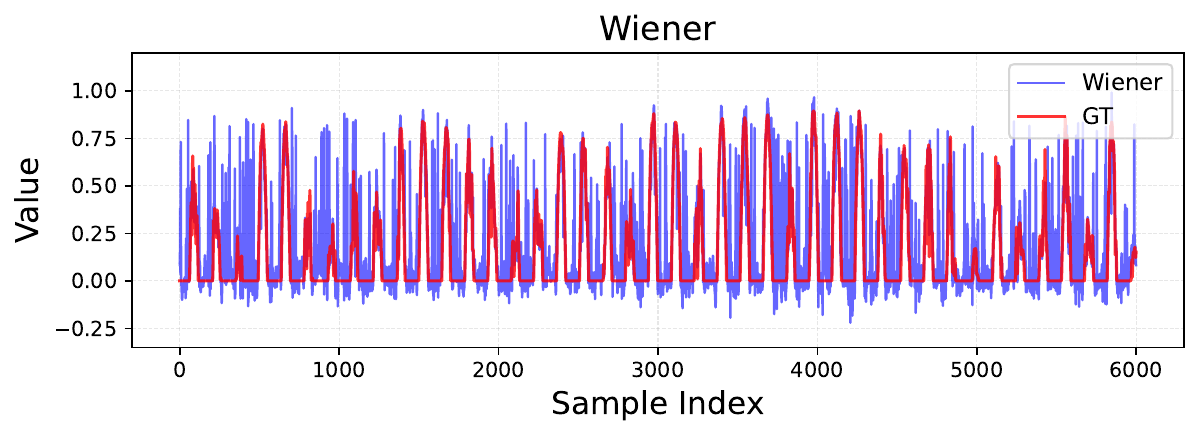}\hspace{0.0em}
\includegraphics[scale=0.4]{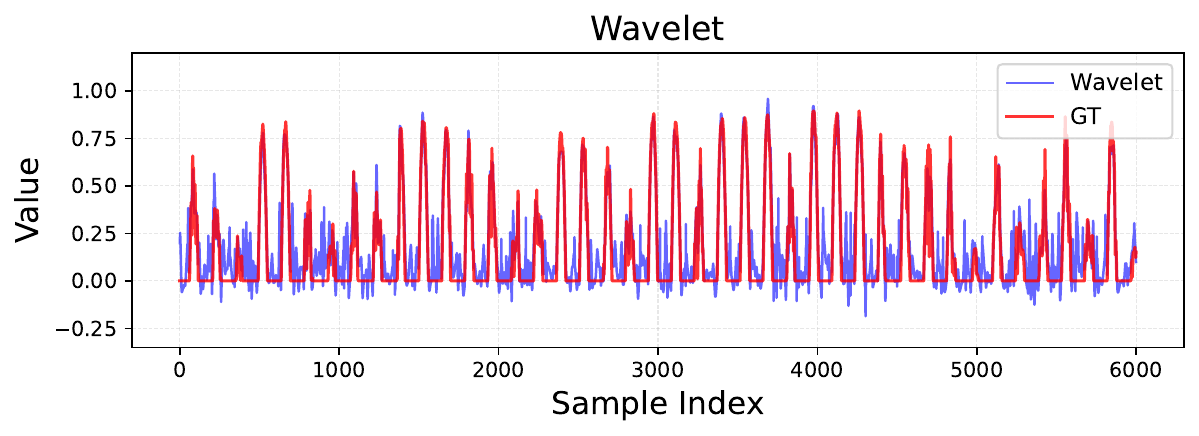}\hspace{0.0em}
\includegraphics[scale=0.4]{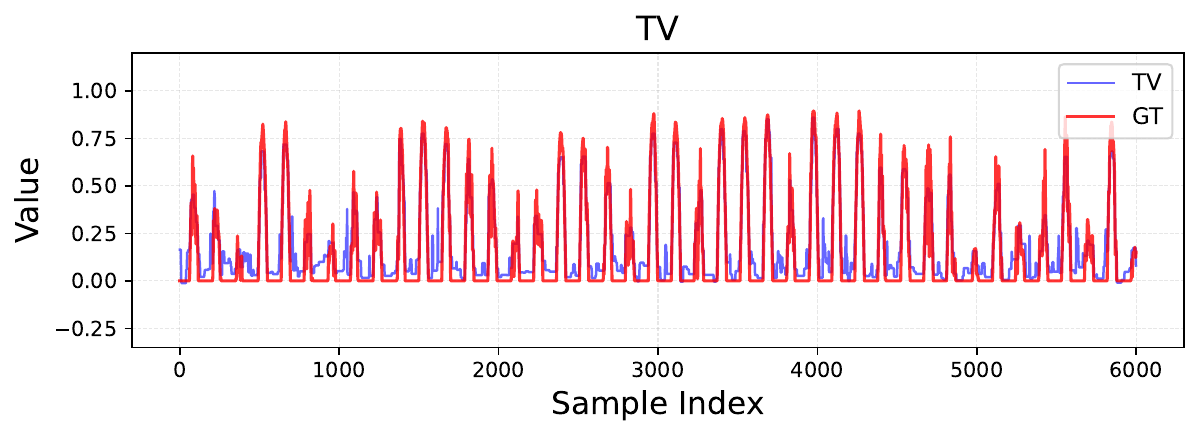}\hspace{0.0em}
\includegraphics[scale=0.4]{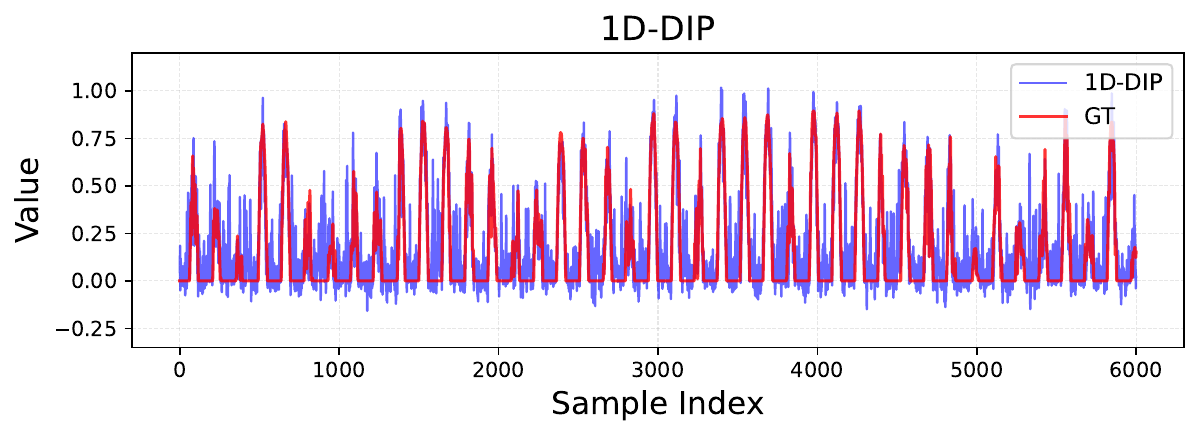}\hspace{0.0em}
\includegraphics[scale=0.4]{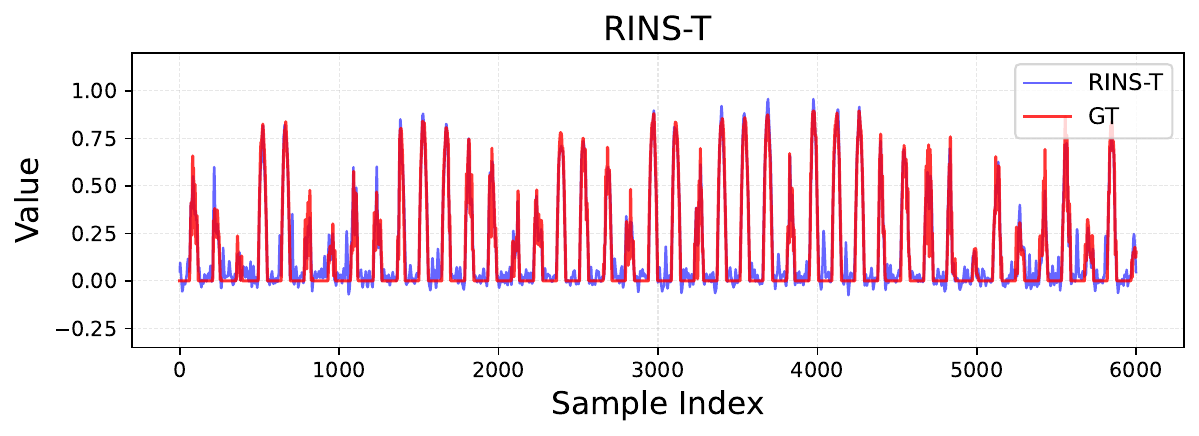}\hspace{0.0em}
\caption{Visual comparison of denoising results for time series data. 
The figure shows the ground truth signal (GT), its noisy counterpart, and the outputs of different denoising methods including Gaussian, Median, Wiener, Wavelet, TV, 1D-DIP, and the proposed RINS-T. This visualization highlights how each method reconstructs the underlying signal structure, with RINS-T providing superior preservation of temporal patterns while effectively reducing noise.}
    \label{fig:main}
\end{figure*}

In contrast, RINS-T  is less effective  on the Audio dataset, where signals are characterized by rapid transients, high-frequency components, and non-stationary behavior. These feature are better captured by Wavelet Denoising, which explicitly promotes sparsity in the time-frequency domain. As a learned prior without explicit frequency localization, RINS-T lacks this targeted sensitivity and may  over-smooth sharp audio  features.

Similarly, 1D-DIP  consistently underperforms, particularly struggling  with non-Gaussian noise, which leads to substantial drops in accuracy.  These results underscore the effectiveness of the key innovations  in RINS-T, such as the robust data-fitting term, input perturbation, convex output combination, and  guided input strategy. These advancements  significantly enhance  reliability and adaptability of deep prior-based denoising methods  across a range   of challenging conditions.

Figure \ref{denoising_chart} presents a bar chart comparing different denoising methods across three distinct scenarios, illustrating the average SNR achieved for each dataset. It shows that the proposed RINS-T method consistently outperforms all other techniques, achieving the highest SNR values in every scenario. The results indicate that while established methods like TV and Wavelet provide robust performance, particularly in Scenario 1, the proposed approach offers superior and more reliable noise suppression across a variety of challenging conditions, confirming its effectiveness as a state-of-the-art denoising solution.

Figure \ref{fig:main} provides a visual comparison of denoising performance across several baseline methods and the proposed RINS-T framework on a representative time series segment for Scenario 3 of Solar dataset. The ground truth signal exhibits temporal structures with sharp transitions, while the noisy input is heavily contaminated, obscuring the underlying patterns. Traditional methods such as Median and Wiener filtering reduce some noise but fail to adequately recover the fine temporal dynamics, often leaving residual fluctuations. Wavelet denoising improves the reconstruction but tends to introduce artifacts in certain regions. TV demonstrates stronger denoising capability, yet they either oversmooth the signal or suppress important local variations. In contrast, RINS-T effectively preserves the sharp transitions and overall temporal structure while simultaneously achieving strong noise suppression. This visual evidence complements the quantitative results, highlighting RINS-T’s ability to balance noise reduction with structural fidelity in time series denoising tasks.

\subsection{Imputation}
In the imputation experiments, the sensing matrix $A$ is defined as a diagonal binary mask $\mathrm{diag}(m)$, where $m$ denotes  observed ($1$) and missing ($0$) entries.

\textbf{Baselines:} We compared RINS-T to several standard approaches: Zero Imputation (filling missing values with zeros), Mean and Median Imputation (using a window size of 15), Spline Interpolation, 1D-DIP, and ImputeFormer \cite{nie2024imputeformer}.

\textbf{Scenarios:} Two scenarios were designed to evaluate imputation performance. In Scenario 1, 20\% of the data was randomly removed (missing completely at random), with an additional 10\% of the data replaced by outliers randomly drawn from a uniform distribution between 0 and 1. Scenario 2 followed the same setup as Scenario 1 but increased the missing data rate to 50\%, making the task more challenging.

\textbf{Results:} Table \ref{tab:imputation} presents a comparison of the imputation methods across three datasets under two scenarios. RINS-T consistently achieves the best performance, with the lowest RMSE and MAE values and the highest SNR across most datasets and scenarios. Its robust design effectively handles missing data and outliers, leading to substantial improvements in imputation accuracy. ImputeFormer also performs competitively, delivering strong results, particularly in Scenario 1, while 1D-DIP generally ranks next but lags behind both RINS-T and ImputeFormer. Other baseline methods, including Zero, Mean, Median, and Spline, exhibit lower performance across most metrics.

\begin{table*}[t]
    \centering
    \caption{Comparison of Different Imputation Methods across 3 Datasets}
    \resizebox{\linewidth}{!}{
    \begin{sc}
    \begin{tabular}{llcccccccccccc}
        \toprule
        \multirow{2}{*}{Scenarios} & \multirow{2}{*}{Method} & \multicolumn{3}{c}{Audio} & \multicolumn{3}{c}{Electricity} & \multicolumn{3}{c}{Solar} & \multicolumn{3}{c}{\cellcolor{gray!20}Average} \\
        \cmidrule(lr){3-5} \cmidrule(lr){6-8} \cmidrule(lr){9-11} \cmidrule(lr){12-14}
        & & RMSE $\downarrow$ & MAE $\downarrow$ & SNR $\uparrow$ & RMSE $\downarrow$ & MAE $\downarrow$ & SNR $\uparrow$ & RMSE $\downarrow$ & MAE $\downarrow$ & SNR $\uparrow$ & \cellcolor{gray!20}RMSE $\downarrow$ & \cellcolor{gray!20}MAE $\downarrow$ & \cellcolor{gray!20}SNR $\uparrow$ \\
        \midrule
        \multirow{6}{*}{Scenario 1} 
        & Zero & 0.4698 & 0.4255 & 0.16 & 0.3667 & 0.3509 & 0.07 & 0.3511 & 0.2019 & -0.66 & \cellcolor{gray!20}0.3959 & \cellcolor{gray!20}0.3261 & \cellcolor{gray!20}-0.14 \\
        & Mean  & 0.1257 & 0.0683 & 11.62 & 0.1338 & 0.0903 & 8.83 & 0.1855 & 0.1032 & 4.88 & \cellcolor{gray!20}0.1483 & \cellcolor{gray!20}0.0873 & \cellcolor{gray!20}8.44 \\
        & Median & 0.1190 & 0.0566 & 12.09 & 0.1294 & 0.0807 & 9.12 & 0.1728 & \underline{0.0643} & 5.50 & \cellcolor{gray!20}0.1404 & \cellcolor{gray!20}\underline{0.0672} & \cellcolor{gray!20}8.90 \\
        & Spline & 0.1510 & 0.0887 & 10.02 & 0.1390 & 0.0733 & 8.49 & 0.2106 & 0.0941 & 3.78 & \cellcolor{gray!20}0.1669 & \cellcolor{gray!20}0.0854 & \cellcolor{gray!20}7.43 \\
        & 1D-DIP & 0.1070 & 0.0804 & 13.01 & \underline{0.0906} & 0.0699 & \underline{12.21} & 0.1238 & 0.0895 & 8.39 & \cellcolor{gray!20}0.1071 & \cellcolor{gray!20}0.0799 & \cellcolor{gray!20}11.20 \\
        & ImputeFormer & \textbf{0.0493} & \textbf{0.0390} & \textbf{19.74} & 0.0909 & \underline{0.0693} & 12.18 & \underline{0.1180} & 0.0938 & \underline{8.81} & \cellcolor{gray!20}\underline{0.0861} & \cellcolor{gray!20}0.0674 & \cellcolor{gray!20}\underline{13.58} \\
        & RINS-T & \underline{0.0585} & \underline{0.0455} & \underline{18.26} & \textbf{0.0831} & \textbf{0.0621} & \textbf{12.96} & \textbf{0.0663} & \textbf{0.0340} & \textbf{13.82} & \cellcolor{gray!20}\textbf{0.0693} & \cellcolor{gray!20}\textbf{0.0472} & \cellcolor{gray!20}\textbf{15.01} \\
        \midrule
        \multirow{6}{*}{Scenario 2} 
        & Zero & 0.4726 & 0.4279 & 0.18 & 0.3669 & 0.3510 & 0.07 & 0.3461 & 0.1982 & -0.64 & \cellcolor{gray!20}0.3952 & \cellcolor{gray!20}0.3257 & \cellcolor{gray!20}-0.13 \\
        & Mean  & 0.1343 & 0.0805 & 11.11 & 0.1433 & 0.0996 & 8.23 & 0.2022 & 0.1278 & 4.03 & \cellcolor{gray!20}0.1599 & \cellcolor{gray!20}0.1026 & \cellcolor{gray!20}7.79 \\
        & Median  & 0.1210 & \underline{0.0597} & 12.02 & 0.1358 & 0.0876 & 8.70 & \underline{0.1793} & \underline{0.0742} & \underline{5.08} & \cellcolor{gray!20}0.1454 & \cellcolor{gray!20}\underline{0.0738} & \cellcolor{gray!20}8.60 \\
        & Spline  & 0.1618 & 0.0972 & 9.49 & 0.1555 & 0.0916 & 7.52 & 0.2318 & 0.1167 & 2.85 & \cellcolor{gray!20}0.1830 & \cellcolor{gray!20}0.1018 & \cellcolor{gray!20}6.62 \\
        & 1D-DIP & 0.1377 & 0.1034 & 10.89 & 0.0933 & 0.0714 & 11.96 & 0.1807 & 0.1214 & 5.01 & \cellcolor{gray!20}0.1372 & \cellcolor{gray!20}0.0987 & \cellcolor{gray!20}9.29 \\
        & ImputeFormer & \textbf{0.0748} & \textbf{0.0576} & \textbf{16.19} & \underline{0.0928} & \underline{0.0708} & \underline{12.01} & 0.1827 & 0.1552 & 4.91 & \cellcolor{gray!20}\underline{0.1168} & \cellcolor{gray!20}0.0945 & \cellcolor{gray!20}\underline{11.04} \\
        & RINS-T & \underline{0.0934} & 0.0723 & \underline{14.27} & \textbf{0.0882} & \textbf{0.0681} & \textbf{12.45} & \textbf{0.0934} & \textbf{0.0484} & \textbf{10.74} & \cellcolor{gray!20}\textbf{0.0917} & \cellcolor{gray!20}\textbf{0.0630} & \cellcolor{gray!20}\textbf{12.49} \\
        \bottomrule
    \end{tabular}
    \end{sc}
    }
    \label{tab:imputation}
\end{table*}

\begin{figure}
    \begin{center}
        \includegraphics[width=\linewidth]{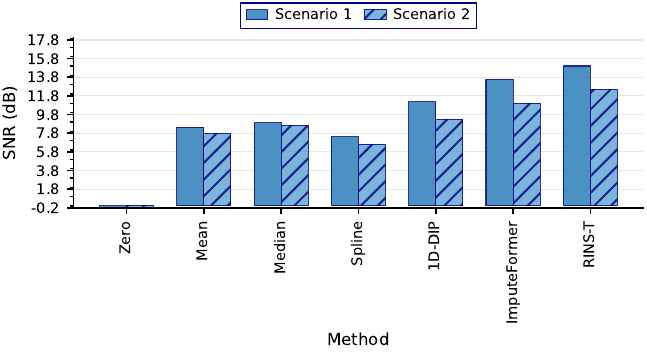}
    \end{center}
    \caption{Comparison of average SNR values across two imputation scenarios using different methods. The proposed RINS-T method consistently achieves the best SNR across all scenarios.} 
    \label{imputation_chart}
\end{figure}

Figure \ref{imputation_chart} presents a bar chart comparing different data imputation methods across two distinct missing-data scenarios, illustrating the average SNR achieved for each dataset. It shows that the proposed RINS-T method significantly outperforms all other techniques, achieving the highest SNR values in both scenarios. The results indicate that while the learning-based 1D-DIP and ImputeFormer methods provide a notable improvement over traditional techniques like mean, median, and spline interpolation, the proposed approach offers superior and more reliable data reconstruction across both scenarios, validating its superiority for imputation.

\subsection{Audio Compressed Sensing}

To further assess the generality and robustness of RINS-T, we conducted compressed sensing (CS) experiments on audio signals. In this setting, the sensing matrix $A$ is a random Gaussian projection, rather than an identity or binary mask. The compression ratio is defined as: 
\begin{equation}
CR = \frac{m}{n},
\end{equation}
where $m$ is the number of observed samples and $n$ is the total number of samples. We evaluated two compression rates 20\% and 50\%, representing high and moderate compression, respectively. Additionally, to test robustness, 10\% of the compressed measurements were intentionally corrupted with outliers.

\textbf{Results:} As reported  in Table~\ref{tab:cs-audio}, RINS-T consistently outperforms 1D-DIP by a substantial  margin across all metrics and compression levels. The model maintains  strong reconstruction performance  even under severe  compression and corruption, demonstrating  that RINS-T effectively addresses a broader range of inverse problems beyond denoising and imputation.

\begin{table}[h]
    \centering
    \caption{Comparison on RINS-T and 1D-DIP for Audio Compressed Sensing}
    \begin{sc}
    \begin{tabular}{ccccc}
        \toprule
        Rate (\%) & Method  & RMSE $\downarrow$ & MAE $\downarrow$ & SNR $\uparrow$ \\
        \midrule
        \multirow{2}{*}{50} & 1D-DIP  & 0.2292 & 0.1804 & 8.33 \\
                            & RINS-T  & \textbf{0.1035} & \textbf{0.0830} & \textbf{15.24} \\
        \midrule
        \multirow{2}{*}{20} & 1D-DIP  & 0.2518 & 0.2053 & 7.52 \\
                            & RINS-T  & \textbf{0.1675} & \textbf{0.1324} & \textbf{11.06} \\
        \bottomrule
    \end{tabular}
    \end{sc}
    \label{tab:cs-audio}
\end{table}

\subsection{Extending to Multivariate Time Series}

While the primary focus of this work is on univariate linear inverse problems in time series, the proposed RINS-T framework can be directly extended to multivariate time series without any architectural modifications. This extension is achieved by  adjusting the input and output dimensions of the deep prior architecture to accommodate multivariate data. To validate RINS-T’s   capability in multivariate settings, we conducted experiments on an electroencephalography (EEG) dataset with 19 channels, evaluating three denoising scenarios. These experiments demonstrate that RINS-T is not restricted  to univariate signals but can effectively process multichannel data. As shown in Table~\ref{tab:eeg_denoising}, RINS-T consistently outperforms classical denoising techniques, including  Gaussian, Median, Wiener, Wavelet, and TV, as well as a neural baseline which is 1D-DIP, across all evaluation metrics (RMSE, MAE, and SNR).

\begin{table}
    \centering
    \caption{Comparison of Different Denoising Methods on Multivariate EEG Dataset}
    \begin{sc}
    \begin{tabular}{llccc}
        \toprule
        \multirow{1}{*}{Scenarios} & \multirow{1}{*}{Method} & RMSE $\downarrow$ & MAE $\downarrow$ & SNR $\uparrow$ \\
        \midrule
        \multirow{8}{*}{Scenario 1}
        & Noisy     & 0.1002 & 0.0798 & 14.37 \\
        & Gaussian  & 0.0494 & 0.0390 & 20.51 \\
        & Median    & 0.0667 & 0.0530 & 17.90 \\
        & Wiener    & 0.0349 & \underline{0.0263} & 23.53 \\
        & Wavelet   & \underline{0.0334} & 0.0265 & \underline{23.92} \\
        & TV        & 0.0343 & 0.0271 & 23.68 \\
        & 1D-DIP    & 0.0465 & 0.0362 & 21.03 \\
        & RINS-T    & \textbf{0.0310} & \textbf{0.0244} & \textbf{24.56} \\
        \midrule
        \multirow{8}{*}{Scenario 2}
        & Noisy     & 0.2731 & 0.2259 & 5.66 \\
        & Gaussian  & 0.0713 & 0.0568 & 17.33 \\
        & Median    & 0.1967 & 0.1574 & 8.52 \\
        & Wiener    & 0.0844 & 0.0661 & 15.86 \\
        & Wavelet   & \underline{0.0696} & \underline{0.0553} & \underline{17.54} \\
        & TV        & 0.0738 & 0.0585 & 17.03 \\
        & 1D-DIP    & 0.0980 & 0.0751 & 14.57 \\
        & RINS-T    & \textbf{0.0666} & \textbf{0.0523} & \textbf{17.92} \\
        \midrule
        \multirow{8}{*}{Scenario 3}
        & Noisy     & 0.1333 & 0.0971 & 11.90 \\
        & Gaussian  & 0.0420 & 0.0319 & 21.92 \\
        & Median    & 0.0780 & 0.0598 & 16.55 \\
        & Wiener    & 0.0529 & 0.0362 & 19.92 \\
        & Wavelet   & 0.0413 & 0.0325 & 22.07 \\
        & TV        & \underline{0.0382} & \underline{0.0302} & \underline{22.75} \\
        & 1D-DIP    & 0.0563 & 0.0432 & 19.37 \\
        & RINS-T    & \textbf{0.0348} & \textbf{0.0270} & \textbf{23.56} \\
        \bottomrule
    \end{tabular}
    \end{sc}
    \label{tab:eeg_denoising}
\end{table}

We further conducted an additional study to confirm that RINS-T effectively considers interdependencies across channels rather than performing independent channel-wise denoising. Specifically, we compared  the denoising performance across multiple target channels (Channels 3, 6, 9, 12, 15, and 18) under two  configurations: (i) using only target channel as input (univariate case), and (ii) using all 19 channels (multivariate case), with identical  noise levels in both settings. As shown  in Table~\ref{tab:multivariate_study}, the denoising performance improves significantly in the multivariate configuration. This result suggests that RINS-T effectively leverages inter-channel correlations, utilizing information across channels rather than relying solely on the temporal dependencies of individual channels.  Thus, the method demonstrates a richer prior beyond channel-wise denoising and is capable of learning dependencies across the multivariate signal.

\begin{table}[h]
    \centering
    \caption{Comparison of Denoising Performance Across Multiple Channels}
    \begin{sc}
    \begin{tabular}{c lccc}
        \toprule
        \centering Channel & Configuration & SNR $\uparrow$ & RMSE $\downarrow$ & MAE $\downarrow$ \\
        \midrule
        \multirow{2}{*}{3}  & Multi Channel  & 25.84 & 0.0267 & 0.0211 \\
                            & Single Channel & 21.78 & 0.0426 & 0.0335 \\
        \midrule
        \multirow{2}{*}{6}  & Multi Channel  & 23.99 & 0.0335 & 0.0273 \\
                            & Single Channel & 22.68 & 0.0389 & 0.0314 \\
        \midrule
        \multirow{2}{*}{9}  & Multi Channel  & 26.37 & 0.0250 & 0.0192 \\
                            & Single Channel & 22.71 & 0.0382 & 0.0301 \\
        \midrule
        \multirow{2}{*}{12} & Multi Channel  & 24.94 & 0.0298 & 0.0238 \\
                            & Single Channel & 22.49 & 0.0394 & 0.0318 \\
        \midrule
        \multirow{2}{*}{15} & Multi Channel  & 26.12 & 0.0259 & 0.0208 \\
                            & Single Channel & 22.76 & 0.0382 & 0.0291 \\
        \midrule
        \multirow{2}{*}{18} & Multi Channel  & 24.64 & 0.0308 & 0.0252 \\
                            & Single Channel & 21.54 & 0.0441 & 0.0344 \\
        \bottomrule
    \end{tabular}
    \end{sc}
    \label{tab:multivariate_study}
\end{table}

\subsection{Ablation Studies}
\textbf{Effects of Learning Strategies:} We performed  ablation studies on the Solar dataset using Scenario 3 of the denoising task, which involves  zero-mean Gaussian noise and 10\% outliers.  Table \ref{tab:ablation_study} shows that each learning strategy improves performance, as evidenced  by increased SNR and decreased  RMSE and MAE. Notably, replacing guided input  with random initialization leads to a substantial drop in performance, underscoring  the limitations of standard deep prior methods that   map from random noise to observed  signals. Furthermore, incorporating convex output combination and input perturbation at each iteration  further enhances the network's  ability to recover clean signals,  resulting in improved  denoising performance.
\begin{table}
    \centering
    \caption{Ablation Study for Denoising (Scenario 3 - Solar Dataset)}
        \resizebox{\linewidth}{!}{
    \begin{sc}
    \begin{tabular}{lccc}
    \toprule
        Experiment & RMSE $\downarrow$ & MAE $\downarrow$ & SNR $\uparrow$ \\
        \midrule
        w/o Convex Combination & 0.0538 & 0.0317 & 15.44 ($\pm$ 0.15) \\
        w/o Input Perturbation & 0.0534 & 0.0352 & 15.50 ($\pm$ 0.19) \\
        w/o Guided Input & 0.0800 & 0.0568 & 11.99 ($\pm$ 0.47) \\
        \midrule
        RINS-T & \textbf{0.0526} & \textbf{0.0311} & \textbf{15.64 ($\pm$ 0.11)} \\
        \bottomrule
    \end{tabular}
    \end{sc}
    }
    \label{tab:ablation_study}
\end{table}

\begin{figure}
\centering
\subfigure[]{\includegraphics[width=0.48\linewidth]{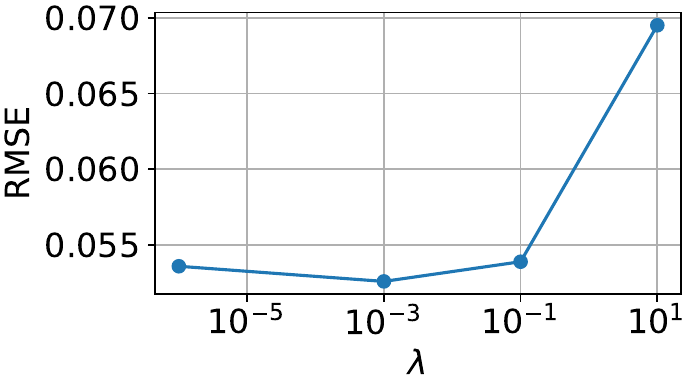}}\label{fig:1a}
\subfigure[] {\includegraphics[width=0.48\linewidth]{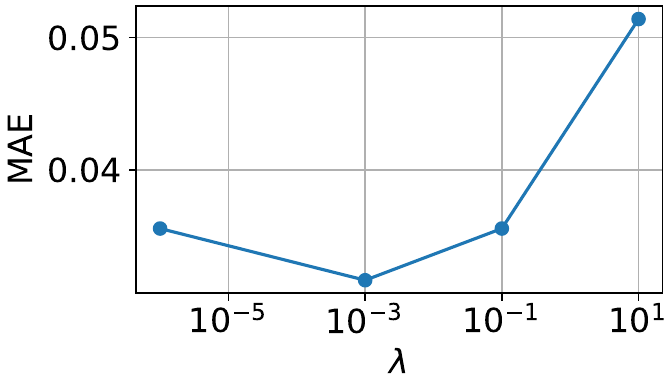}}\label{fig:1b}
\caption{Effect of $\lambda$ on Denoising Task (Scenario 3 - Solar Dataset)} \label{fig:1}
\label{fig:lambda}
\end{figure}

\textbf{Effect of Sparsity Regularization:} To assess the impact of the sparsity regularization term on model robustness, we evaluate the performance of our method under varying values of $\lambda$ on Scenario 3 of the Solar dataset, which contains significant noise and outlier contamination. This experiment aims to understand how different levels of regularization influence the model's ability to handle sparse corruptions. When  $\lambda$ is small, the model's performance remains stable and shows  less sensitivity to  this hyperparameter, yielding  consistent  RMSE and MAE values as illustrated in Figure \ref{fig:lambda}. However, as  $\lambda$ increases, the model's sensitivity to outliers becomes more pronounced, resulting in a noticeable decline in performance. This behavior is intuitive because, in the optimization problem (\ref{eq: inverseprob}), as $\lambda$ becomes very large, the sparsity term in the objective function dominates the minimization process,  driving the sparsity component to zero ($s^*=0$). As a result, the model loses its ability to effectively manage sparse contamination, becoming highly sensitive to outliers. Figure \ref{fig:lambda} clearly demonstrates this trend, showing that small to moderate values of $\lambda$ result in the most consistent and reliable performance across all metrics. Therefore, although a theoretical optimum for $\lambda$ can be derived when noise statistics are known, RINS-T maintains robust performance across a range of small values in practice. For real-world applications where such statistics are typically unavailable, we recommend initializing $\lambda$ with a small value (e.g., $\lambda \leq 0.01$ for normalized data), which provides a reliable and effective default.

\textbf{Effect of Weighting Factor for Convex Combination:} We evaluate the influence of the convex combination weighting factor $\alpha$ to understand its role in stabilizing the iterative update process in RINS-T. We set $\alpha = 0.5$ in all experiments, as it provides a good balance between the previous and current outputs in the update rule. To assess  the sensitivity of RINS-T to this parameter, we performed  additional experiments on Scenario 3 of the Solar dataset (denoising task). The results in Table \ref{tab:alpha} demonstrate  that the model is robust to changes  in $\alpha$, with optimal RMSE performance achieved  at $\alpha = 0.5$. Notably, changes in $\alpha$ lead to only minor fluctuations in both RMSE and MAE, indicating that precise  fine-tuning of this parameter is not necessary for effective performance.

\begin{table}[h]
    \centering
    \caption{Ablation study showing RMSE and MAE for different values of $\alpha$ on Scenario 3 of the Solar dataset.}
      \resizebox{\linewidth}{!}{
    \begin{tabular}{lcccccccccc}
        \toprule
        \textbf{$\alpha$} & \multicolumn{2}{c}{0.1} & \multicolumn{2}{c}{0.3} & \multicolumn{2}{c}{0.5} & \multicolumn{2}{c}{0.7} & \multicolumn{2}{c}{0.9} \\
        \cmidrule(r){2-3} \cmidrule(r){4-5} \cmidrule(r){6-7} \cmidrule(r){8-9} \cmidrule(r){10-11}
         & RMSE & MAE & RMSE & MAE & RMSE & MAE & RMSE & MAE & RMSE & MAE \\
        \midrule
         & 0.0534 & 0.0313 & 0.0527 & \textbf{0.0308} & \textbf{0.0526} & 0.0311 & 0.0531 & 0.0323 & 0.0536 & \textbf{0.0308} \\
        \bottomrule
    \end{tabular}
    }
    \label{tab:alpha}
\end{table}

\begin{figure}
\centering
\subfigure[]{\includegraphics[width=0.48\linewidth]{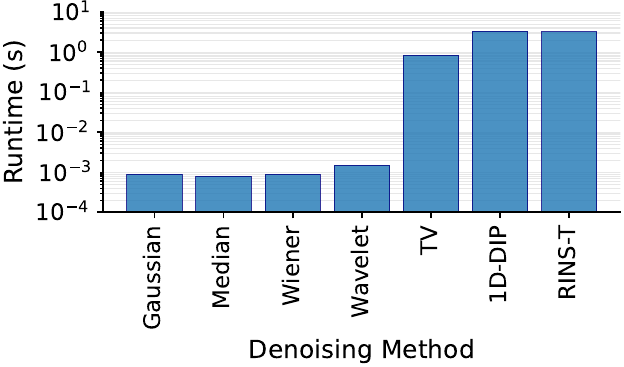}}\label{fig:1a}
\subfigure[] {\includegraphics[width=0.48\linewidth]{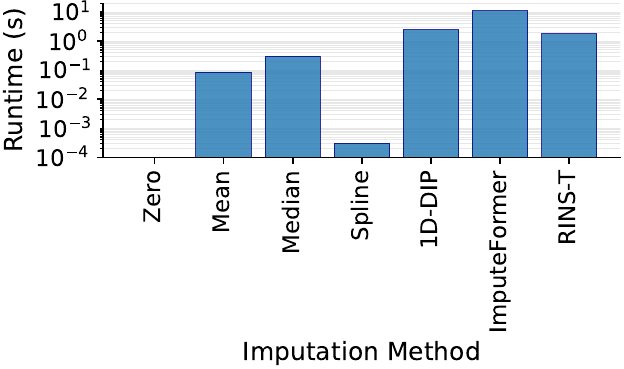}}\label{fig:1b}
\caption{Comparison of Running Times for (a) Denoising and (b) Imputation Methods} \label{fig:1}
\label{fig:runningtime}
\end{figure}

\textbf{Running Time:} Figure \ref{fig:runningtime} compares the running times of various denoising and imputation methods on the NVIDIA GeForce RTX 2080 Ti GPU and Intel Xeon E5-2620 v4 CPU for the Electricity dataset. For denoising (Scenario 3), filtering-based methods such as Gaussian, Median, and Wiener are the fastest. In contrast, TV, 1D-DIP, and RINS-T require more time, although RINS-T converges faster than 1D-DIP due to its guided input. For imputation (Scenario 2), Zero imputation is the fastest, followed by Spline, Mean, and Median methods. Deep learning approaches, including 1D-DIP, ImputeFormer, and RINS-T, require more computational time; among them, RINS-T achieves the shortest runtime.

\subsection{Pointwise Denoising or Sequence-wise Denoising?}

We provide evidence that the model leverages temporal structure in the data, moving beyond simple pointwise denoising. Consider a 1D time series $x = [x_1, x_2, ..., x_T]$ and a 1D convolutional filter $w = [w_1, w_2, ..., w_k]$ of length $k$. The 1D convolution operation produces an output $y$, where each element $y_t$ is computed as:

\begin{equation}
y_t = \sum_{i=1}^{k} w_i \cdot x_{t - i + 1} = w_1 x_t + w_2 x_{t-1} + \dots + w_k x_{t-k+1}
\end{equation}

Each output $y_t$ is thus a weighted sum of a contiguous window of $k$ time steps from the input, directly incorporating temporal dependencies from multiple time points.

To empirically verify that our model leverages these temporal dependencies, we conducted an experiment on Scenario 3 of the Electricity dataset. Specifically, we disrupted the temporal structure of the input  by randomly permuting the time indices. If the model were limited to  pointwise denoising, this manipulation would  have minimal affect on its performance. However, as shown in Table~\ref{tab:temporal-dependencies}, the model's performance degrades significantly  when temporal dependencies are removed, confirming its reliance on temporal context.

\begin{table}[h]
    \centering
    \caption{Effect of Temporal Dependency Manipulation on Model Performance (Electricity Dataset, Scenario 3)}
    \begin{sc}
    \begin{tabular}{lcc}
        \toprule
        Condition & RMSE $\downarrow$ & MAE $\downarrow$ \\
        \midrule
        w Permutation & 0.0913 & 0.0701 \\
        w/o Permutation & \textbf{0.0698} & \textbf{0.0536} \\
        \bottomrule
    \end{tabular}
    \end{sc}
    \label{tab:temporal-dependencies}
\end{table}

\section{Conclusion}
In this work, we propose RINS-T, a deep prior framework that leverages the Huber loss as its data-fitting term. The Huber loss naturally emerges  from two complementary perspectives: (1) as the solution to a convex optimization problem with $\ell_1$-norm sparsity constraints, and (2) from a probabilistic viewpoint   that blends  Gaussian and Laplace noise models. This dual foundation  provides strong theoretical justification for  the Huber loss’s  robustness to outliers and contaminated noise. To further improve optimization stability and performance, we augment our framework with guided input initialization, input perturbation, and convex output combination. By leveraging the intrinsic one-dimensional structure of time series data, RINS-T achieves consistent performance improvements across diverse datasets. However, we observed that RINS-T can oversmooth sharp audio signals, leading to underrepresentation of high-frequency or transient features. Exploring architectural extensions, such as wavelet-inspired layers or multi-scale filters, is a promising direction for addressing this limitation. Future work may therefore investigate both these architectural refinements and further theoretical links between noise models and loss functions, as well as extend the framework to nonlinear inverse problems where the forward model exhibits nonlinear relationships, potentially paving the way for advanced robust estimation methods in related domains.

\bibliographystyle{IEEEtran}
\bibliography{IEEEabrv,main}

\end{document}